\documentclass[letterpaper]{article} 
\usepackage[]{aaai2026}  
\usepackage{times}  
\usepackage{helvet}  
\usepackage{courier}  
\usepackage[hyphens]{url}  
\usepackage{graphicx} 
\usepackage{natbib}  
\usepackage{caption} 
\usepackage{algorithm}
\usepackage{algorithmic}

\usepackage{newfloat}
\usepackage{listings}
\DeclareCaptionStyle{ruled}{labelfont=normalfont,labelsep=colon,strut=off} 
\floatstyle{ruled}
\newfloat{listing}{tb}{lst}{}
\floatname{listing}{Listing}
\PassOptionsToPackage{table}{xcolor}  
\usepackage[dvipsnames]{xcolor}                  
\usepackage{colortbl}                 
\usepackage{graphicx}

\usepackage{amssymb}
\usepackage{amsmath} 
\usepackage{booktabs} 
\usepackage{adjustbox}
\usepackage{enumitem} 
\definecolor{DarkTurquoise}{RGB}{0, 206, 209}
\definecolor{DeepTeal}{RGB}{0, 102, 102}
\definecolor{Teal}{RGB}{0, 128, 128}
\definecolor{DarkCyan}{RGB}{0, 139, 139}
\definecolor{DarkBurntOrange}{RGB}{183, 65, 14}

\title{Adaptive Agent Selection and Interaction Network for Image-to-point cloud Registration
}
\author {
    Zhixin Cheng\textsuperscript{\rm }$^{1,2}$,
    Xiaotian Yin\textsuperscript{\rm }$^{3}$,
    Jiacheng Deng\textsuperscript{\rm }$^{1,2}$,
    Bohao Liao\textsuperscript{\rm }$^{1,2}$,
    Yujia Chen\textsuperscript{\rm }$^{3}$,
    Xu Zhou\textsuperscript{\rm }$^{4}$,
    Baoqun Yin\textsuperscript{\rm }$^{1}$,
    Tianzhu Zhang\textsuperscript{\rm }$^{1,2}$
    \thanks{Corresponding author.}
}
\affiliations {
    \textsuperscript{\rm }$^1$School of Information Science and Technology, University of Science and Technology of China \\
$^2$National Key Laboratoray of Deep Space Exploration, Deep Space Exploration Laboratory \\
$^3$Institute of Advanced Technology, University of Science and Technology of China \\
$^4$Sangfor Technologies \\
    {chengzhixin, xiaotianyin, dengjc, liaobh, yujia\_chen}@mail.ustc.edu.cn, zhouxu@sangfor.com.cn,  {bqyin, tzzhang}@ustc.edu.cn

}

\usepackage{bibentry}

\begin{document}

\maketitle

\begin{abstract}
Typical detection-free methods for image-to-point cloud registration leverage transformer-based architectures to aggregate cross-modal features and establish correspondences. However, they often struggle under challenging conditions, where noise disrupts similarity computation and leads to incorrect correspondences. Moreover, without dedicated designs, it remains difficult to effectively select informative and correlated representations across modalities, thereby limiting the robustness and accuracy of registration.
To address these challenges, we propose a novel cross-modal registration framework composed of two key modules: the Iterative Agents Selection (IAS) module and the Reliable Agents Interaction (RAI) module. IAS enhances structural feature awareness with phase maps and employs reinforcement learning principles to efficiently select reliable agents. RAI then leverages these selected agents to guide cross-modal interactions, effectively reducing mismatches and improving overall robustness.
Extensive experiments on the RGB-D Scenes v2 and 7-Scenes benchmarks demonstrate that our method consistently achieves state-of-the-art performance.
\end{abstract}


\section{Introduction}
Image-to-point cloud registration (I2P) aims to estimate the rigid transformation aligning the point cloud with the camera coordinate system, given an image and a point cloud from the same scene. This typically involves cross-modal feature matching to establish correspondences, followed by pose estimation to compute the rotation and translation matrices. As a crucial step in various vision tasks, such as 3D reconstruction \cite{3dreconstruction,deng2024unsupervised}, SLAM \cite{slam,denghierarchical}, and visual localization \cite{visuallocalization,diffreg}, I2P plays a vital role. However, there is a significant gap between the two modalities: images are dense, structured 2D grids, while point clouds consist of sparse, unordered, and irregular 3D data. Effectively bridging this gap to improve feature aggregation across these modalities remains a key challenge in cross-modal registration.

\begin{figure}[!t]
\centering
\includegraphics[width=\columnwidth]{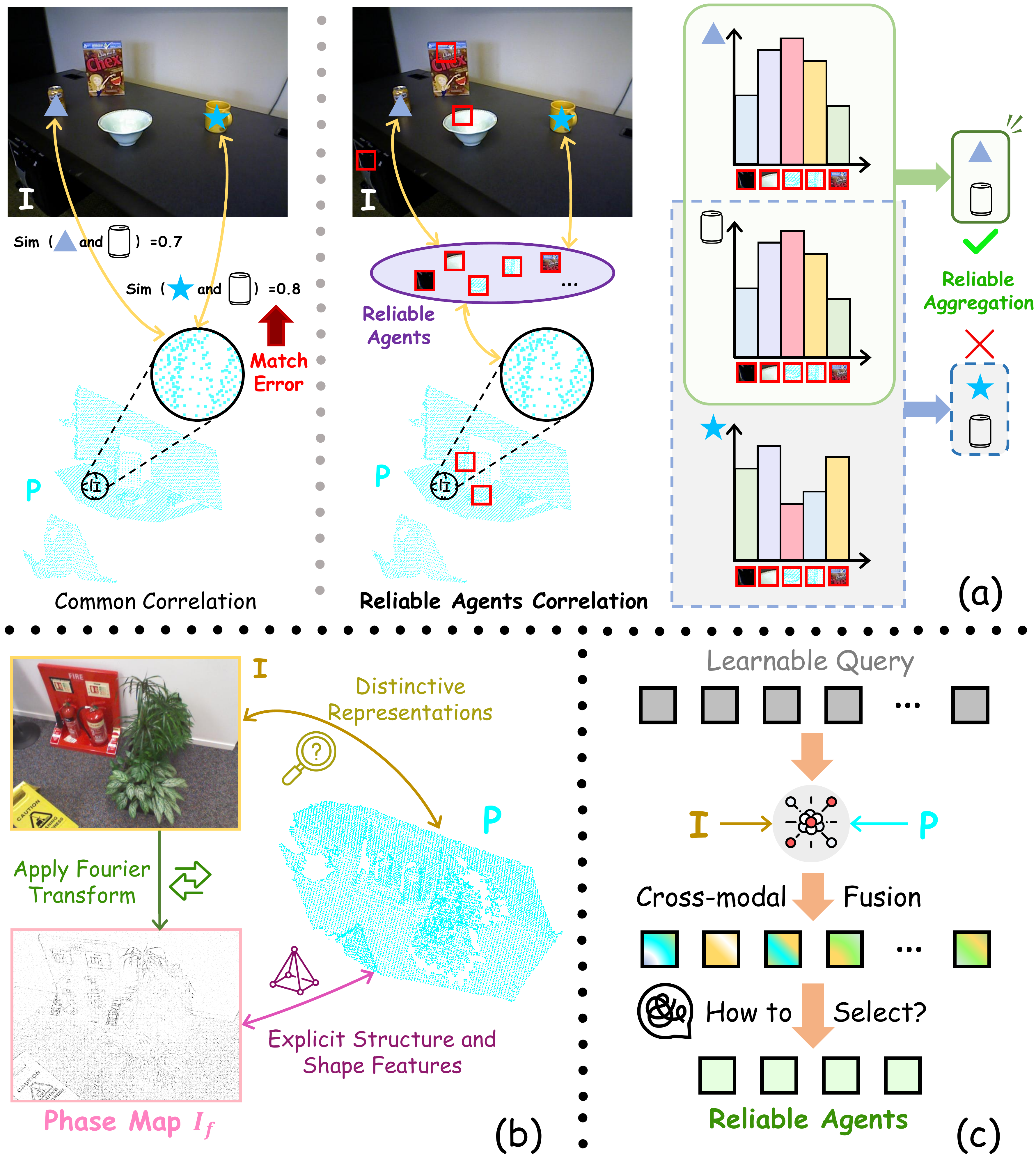} 
\caption{(a) Visual comparison of common and reliable agents correlations
(b) Visualization of phase map enhancement.
(c) Procedure and challenges of the reliable agents interaction method.}
\label{fig1}
\end{figure}

Current solutions to I2P challenges generally fall into two categories. Recently, detection-free approaches \cite{cofii2p, matr2d3d} have emerged as the mainstream, gradually replacing traditional detect-then-match methods \cite{2d3dmatchnet, p2, corri2p}. The detect-then-match paradigm involves independently detecting 2D keypoints in images and 3D keypoints in point clouds, and then matching them based on semantic features. However, this approach has two major limitations: first, keypoint detection is inherently modality dependent, image keypoints rely on texture and color, while point cloud keypoints depend on local geometry—making it challenging to extract consistent keypoints. Second, the significant differences between 2D and 3D descriptors impedes effective cross-modal alignment. Thus, 2D3D-MATR \cite{matr2d3d} introduces the first coarse-to-fine detection-free framework. This framework extracts features from both modalities, performs coarse patch-level matching, and then refines these matches into fine-grained pixel-to-point correspondences. Finally, it estimates rigid transformations using PnP-RANSAC \cite{pnp,ransac}. However, in the scenarios with repetitive structures \cite{repetitive}, non-overlapping regions, or varying illumination, directly aggregating features through a conventional transformer \cite{transformer} may introduce substantial noise, thereby degrading matching accuracy \cite{agent} and ultimately affecting overall registration performance.

Building on the discussion of detection-free methods, we identify two critical issues that are essential for accurate and robust registration between images and point clouds.
\textbf{Firstly, we need to find effective ways to aggregate informative features from both modalities to reduce cross-modal mismatches.} Traditional transformer-based approaches often compute similarities directly between image and point cloud features. However, in scenes with repetitive structures or non-overlapping regions, such similarity measures can lead to incorrect correspondences. For example, as illustrated in Figure 1(a), a soda can might be confused with a structurally similar object. By aggregating features based on key informative regions, we could reduce such ambiguities and improve registration accuracy.
\textbf{Therefore, it is crucial to reliably select representative features from both modalities to enhance registration quality.} As images mainly capture texture and point clouds capture geometry, modality-specific encoding often leads to distinct feature representations. To bridge this gap, we introduce phase maps to enhance the image’s sensitivity to structural edges, thereby facilitating the selection of representative features (Figure 1(b)). Building on this, we further explore learning-based queries to aggregate features across modalities (Figure 1(c)). However, without dedicated design, the full potential of the model for cross-modal registration may not be realized, as identifying optimal and reliable fusion agents is essential.

In response to these concerns, we propose Adaptive Agent Selection and Interaction Network (A2SI) for image-to-point cloud registration. Our framework consists of two main modules: Iterative Agent Selection (IAS) module and Reliable Agent Interaction (RAI) module.
In the IAS, we extract phase maps from images to highlight structural edge information, which helps reduce the domain gap between image and point cloud features. 
Considering that reinforcement learning \cite{reinforcement} supports reward-guided scoring and selection, we adapt this idea to our task and propose a lightweight Tri-Stage Agents Optimization Strategy, designed to identify reliable agents that most enhance the model’s registration capability.
In the RAI module, we utilize these reliable agents to efficiently aggregate cross-modal features, moving away from traditional transformer-based fusion methods. The RAI module greatly minimizes the noise caused by repetitive structures and non-overlapping regions, resulting in higher-quality matching with lower computational costs.

Our main contributions are summarized as follows:

\begin{itemize}[leftmargin=*, itemsep=1pt, topsep=1pt, parsep=0.5pt]
    \item We propose a novel Adaptive Agent Selection and Interaction Network for image-to-point cloud registration, which achieves excellent accuracy and strong generalization ability.
    \item We design the Iterative Agents Selection module, which leverages phase maps to enhance structural awareness in image features, and a Tri-Stage Agents Optimization Strateg selects reliable agents containing key information. In the Reliable Agents Interaction module, we replace traditional transformer fusion with agent-guided interaction for improved matching and reduced errors.
    \item Extensive experiments and ablation studies on two benchmarks, RGB-D Scenes v2 and 7-Scenes, demonstrate the superiority of our method, establishing it as a new state-of-the-art for image-to-point cloud registration.
\end{itemize}

\section{Related works}
In this section, we briefly overview related works on I2P registration, including stereo image registration, point cloud registration, and inter-modality registration. 
\subsection{Image Registration}
Early image registration methods relied on handcrafted keypoints and descriptors, such as SIFT~\cite{sift} and ORB~\cite{orb}, to establish correspondences. Later, learning-based approaches like SuperGlue~\cite{superglue}, which leverage transformers~\cite{transformer}, improved robustness but still struggled in textureless regions due to their reliance on keypoint detection. To overcome this limitation, detector-free methods such as LoFTR~\cite{loftr} and Efficient LoFTR~\cite{efficientloftr} adopt coarse-to-fine matching strategies combined with global attention, enabling dense correspondence estimation without explicit keypoints.

\subsection{Point Cloud Registration}
Point cloud registration has progressed from handcrafted descriptors, such as PPF~\cite{ppf} and FPFH~\cite{fpfh}, to deep learning-based methods. CoFiNet~\cite{cofinet} introduces a detector-free coarse-to-fine pipeline, while GeoTransformer~\cite{geotransformer} further advances performance by leveraging transformers to model global geometry and replacing RANSAC~\cite{ransac} with a local-to-global registration framework.

\begin{figure*}[t]
\centering
\includegraphics[width=\textwidth]{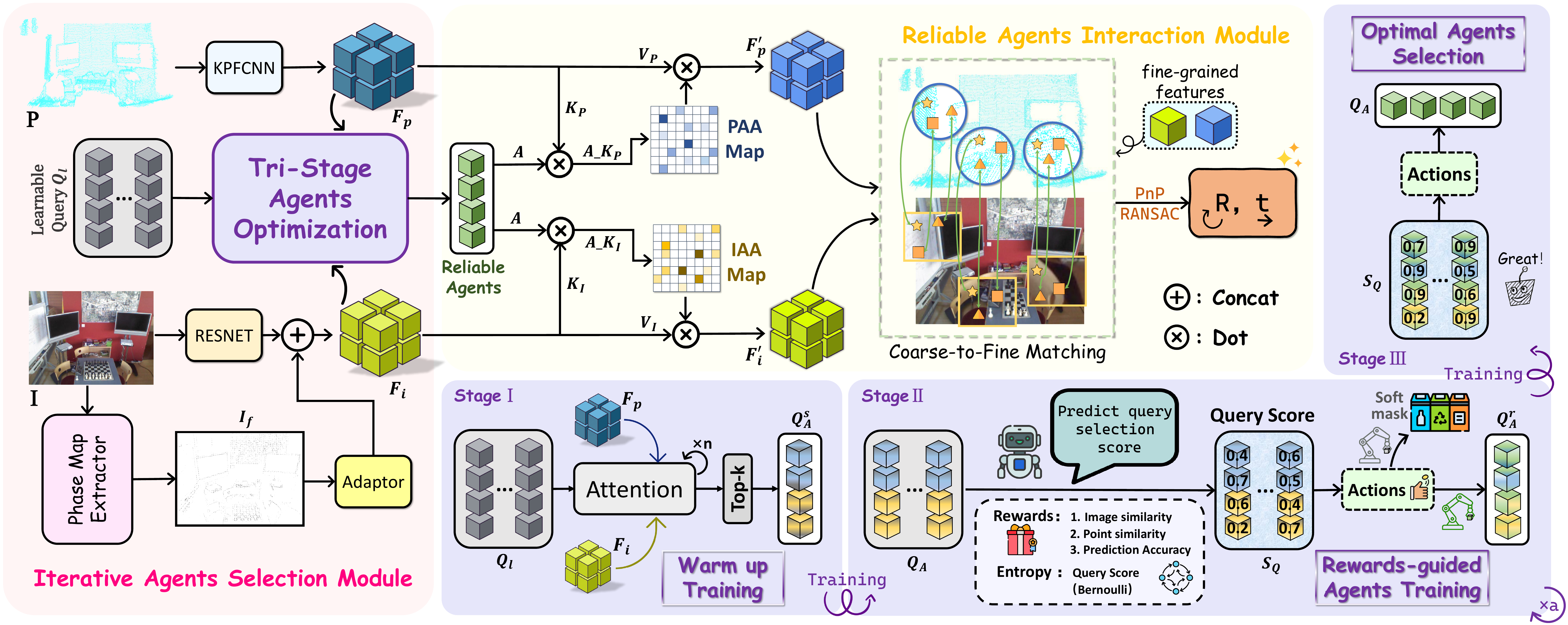} 
\caption{Overall pipeline of A2SI.
A phase map extractor first enhances image feature sensitivity to edge structures. The Tri-Stage Agents Optimization strategy then selects informative reliable agents from redundant learnable queries. These agents guide cross-modal aggregation in the Reliable Agents Interaction Module, progressively establishing dense correspondences in a coarse-to-fine manner. Finally, a rigid transformation is estimated via PnP with RANSAC. }
\label{fig2}
\end{figure*}

\subsection{Inter-Modality Registration}
Compared to intra-modality registration, inter-modality registration presents greater challenges due to significant domain discrepancies. Traditional approaches typically follow a detect-then-match paradigm. For instance, 2D3DMatch-Net~\cite{2d3dmatchnet} leverages SIFT~\cite{sift} for keypoint detection, while ISS~\cite{iss} extracts local patches from point clouds and employs CNNs or PointNet~\cite{pointnet} for descriptor learning and matching. P2-Net~\cite{p2} improves efficiency by jointly detecting and matching points in a single step. However, in cross-modal scenarios, keypoint detection accuracy often degrades~\cite{keypoint,keypoint2}, limiting performance and motivating the rise of detector-free methods.  
2D3D-MATR~\cite{matr2d3d} employs a coarse-to-fine strategy, using a transformer for coarse matching and refining the results with PnP-RANSAC, thus eliminating keypoint detection and improving descriptor alignment, while B2-3D~\cite{cheng11} further enhances accuracy through uncertainty modeling and domain adaptation.  
Building on 2D3D-MATR, our proposed A2SI also adopts a detector-free design, focusing on selecting features that capture region information for cross-modal aggregation, thereby setting a new benchmark for image-to-point cloud registration.

\section{Method}
Let \( I \in \mathbb{R}^{H \times W \times 3} \) represent an image and \( P \in \mathbb{R}^{N \times 3} \) represent a point cloud from the same scene. Here, \( H \) and \( W \) denote the height and width of the image, respectively, while \( N \) is the number of 3D points in the point cloud. The objective of image-to-point cloud registration is to estimate a rigid transformation \([R, t]\), where the rotation matrix \( R \in \mathrm{SO}(3) \) and the translation vector \( t \in \mathbb{R}^3 \).

We propose A2SI, a detection-free registration framework that efficiently aggregates cross-modal features. 
The Iterative Agents Selection module incorporates phase information to enhance edge sensitivity.
Additionally, a lightweight Tri-Stage Agents Optimization strategy, inspired by reinforcement learning, is employed to select reliable agents.
These are refined through a Reliable Agents Interaction Module, improving transformer-based coarse matching. Dense correspondences are further refined using high-resolution features, and the final rigid transformation is estimated with PnP + RANSAC.

\subsection{Iterative Agents Selection Module}
We adopt ResNet \cite{resnet} with an FPN \cite{fpn} and KPFCNN \cite{kpfcnn} to extract 2D and 3D features with positional encoding, respectively. 
Images and point clouds use fundamentally different encoding methods based on texture and structure, respectively. This leads to feature discrepancies that hinder effective fusion and reduce registration accuracy.

\subsubsection{Phase Map Extractor.}
In the frequency domain, previous studies \cite{fourier} have shown that the phase component of the Fourier spectrum retains high-level statistical information, which plays a critical role in preserving the structural and textural characteristics of images. Inspired by this, we propose to leverage Fourier phase to enhance the feature correlation between images and point clouds. Specifically, we design a phase map extractor.

Given an image \( x \in \mathbb{R}^{H \times W \times 3} \), we apply a two-dimensional Fourier transform \( \mathcal{F}(x) \) as:
\begin{equation}
    \mathcal{F}(x)_{u,v} = \sum_{i=0}^{H-1} \sum_{j=0}^{W-1} x_{i,j} \cdot e^{-J2\pi \left( \frac{ui}{H} + \frac{vj}{W} \right)}\text{,}
\end{equation}
where \( J \) refers to the imaginary unit, and \( u \) and \( v \) denote the frequency coordinates along the horizontal and vertical axes, respectively. Then we can acquire the corresponding amplitude \( \mathcal{B} \) and phase \( \Phi \) as:
\begin{equation}
    \mathcal{B}(x)_{u,v} = \left| \mathcal{F}(x)_{u,v} \right|,
\end{equation}
\begin{equation}
    \Phi(x)_{u,v} = \arg\left(\mathcal{F}(x)_{u,v}\right) = \arctan\left( \frac{\text{Im}(\mathcal{F}(x)_{u,v})}{\text{Re}(\mathcal{F}(x)_{u,v})} \right),
\end{equation}
where \( \text{Im}(\cdot) \) and \( \text{Re}(\cdot) \) denote the imaginary and real parts of \( \mathcal{F}(x) \), respectively.

To generate the final phase map, we fix the amplitude to an average constant \( c^x \) and apply an inverse Fourier transform to obtain the phase texture map \( {I_f}^x \):
\begin{equation}
    {I_f}^x = \mathcal{F}^{-1} \left( \Phi(x)_{u,v} \cdot e^{-Jc^x} \right),
\end{equation}
where \( \mathcal{F}^{-1} \) denotes the inverse Fourier transform.

Then, we extract the phase representation \( \phi \) using a three-layer lightweight CNN adaptor and fuse it with the original image features to obtain the coarse-grained image representation \( F_i \in \mathbb{R}^{H \times W \times C} \). The corresponding coarse-grained point cloud feature is denoted as \( F_p \in \mathbb{R}^{N \times C} \).

\subsubsection{Tri-Stage Agents Optimization Strategy.}
To address challenges such as repetitive structures, non-overlapping regions, and variations in lighting, we aim to effectively combine image and point cloud features by selecting reliable agents that capture correlations across modalities (see Figure 3). Previous studies \cite{agent} typically rely on a fixed number of initialized agents, which limits adaptability and may hinder optimal feature representation. While reinforcement learning can facilitate the feedback-driven selection of representative and dependable features, an over-reliance on existing representations may impede model performance.

To overcome these limitations, we initialize redundant learnable queries and propose a Tri-Stage Agents Optimization Strategy. This approach allows us to adaptively select the most reliable agents.
\paragraph{Stage \uppercase\expandafter{\romannumeral 1}: 
 Warm up Training.} We initialize a redundant set of learnable queries, denoted as $Q_l$. These queries interact with image and point cloud features through \( n \) layers of attention, and the resulting aggregated representations are referred to as \( Q_A \) (not explicitly shown in the Stage \uppercase\expandafter{\romannumeral 1} illustration).
Each query is associated with a learnable query score, which is maintained in a dedicated query evaluation block, allowing the model to weigh the importance of each query. We then select the top-$k$ queries based on their query score and refer to them as $Q_A^s$ for subsequent aggregation. The main objective of this stage is to enable the queries $Q_l$ to quickly learn meaningful cross-modal representations, thereby facilitating convergence and establishing a solid foundation for the later stages.

\paragraph{Stage \uppercase\expandafter{\romannumeral 2}: Rewards-guided Agents Training.} We aim to identify a subset of reliable agents  that contribute most to performance improvement.
To this end, we analyze the quality of each query from both local and global perspectives, rather than selecting them based on a top-$k$ criterion.
At the local level, we evaluate each query based on its similarity to image and point cloud features. Let $F_q$ be the feature of a query, and $F_i$, $F_p$ be the image and point cloud features, respectively. The local reward is defined as:
\begin{equation}
    \text{\textit{Local\_Reward}} = \frac{1}{2} \left[ \cos(F_q, F_i) + \cos(F_q, F_p) \right].
\end{equation}

At the global level, we assess the contribution of each query by how much it helps reduce the overall task loss. We derive the $L_{rl}$ from the the main task loss and define the global reward as the inverse of $L_{rl}$:
\begin{equation}
    \text{\textit{Global\_Reward}} = 1.0 / \mathcal{L}_{rl},
\end{equation}
where lower task loss indicates better matching and thus higher global reward.
To balance these two types of rewards, we introduce a dynamic weighting strategy controlled by a fusion coefficient $\alpha$ that evolves over time. Specifically, in the early period of Stage II, the model has not yet converged, and global signals offer more informative gradients. As training progresses, we gradually increase the importance of the local reward to encourage convergence and prevent overfitting. 
We define the fusion coefficient as $\alpha = 1 - e^{-\textit{epoch} / \tau}$, where $\textit{epoch}$ is the current training iteration and $\tau$ is a predefined time constant that controls the growth rate of $\alpha$.  
To enhance adaptability and avoid premature convergence, $\tau$ is progressively decayed during training with a lower bound, allowing $\alpha$ to gradually increase and adjust the balance between local and global rewards.  
The final reward is computed as a weighted combination of the local and global rewards using this fusion coefficient.
\begin{equation}
    \text{\textit{reward}} = \alpha \cdot \text{\textit{Local\_Reward}} + (1 - \alpha) \cdot \text{\textit{Global\_Reward}}.
\end{equation}

Based on the learned score \( S_q \) for each query, we apply a sigmoid function to normalize it to the range \([0, 1]\). We then perform Bernoulli sampling to determine which k queries are selected. The queries that are positively sampled form the set \( Q_A^r \), which will engage in cross-modal interaction and forward propagation.

To stabilize training and maintain gradient flow from the unselected queries, we employ a soft masking strategy. In this approach, each query is assigned a \textit{soft\_mask} defined as follows:
\begin{equation}
    \text{\textit{soft\_mask}} = \beta + (1 - \beta) \cdot a,
\end{equation}
where $a \in \{0, 1\}$ is the Bernoulli action, and $\beta$ is a predefined minimum weight ensuring unselected queries retain a small influence.
We then enhance query selection using the reinforce algorithm, with the log-probability for each query computed as:
\begin{equation}
    \log P(a) = a \cdot \log(p) + (1 - a) \cdot \log(1 - p),
\end{equation}
where $p = \sigma(S_q)$ is the sampling probability. The reinforcement loss is:
\begin{equation}
    \mathcal{L}_{g} = - \left( \text{\textit{reward}} - \overline{\textit{reward}} \right) \cdot \log P(a),
\end{equation}
where $\overline{\textit{reward}}$ denotes the mean of the rewards across the sampled queries, which serves as a baseline to reduce variance. The resulting loss is used to update the query scores \( S_q \) through backpropagation.

To encourage exploration and avoid premature convergence of the query selection policy, we incorporate an entropy regularization term on the query sampling probabilities. This entropy bonus promotes diversity by penalizing over-confident selections. The entropy for each query is defined as:
\begin{equation}
\text{\textit{Entropy}} = - \left[ p \cdot \log(p) + (1 - p) \cdot \log(1 - p) \right],
\end{equation}
the final Stage \uppercase\expandafter{\romannumeral 2} loss with entropy regularization becomes:
\begin{equation}
\mathcal{L}_{full} = \mathcal{L}_{g} - \mu \cdot \sum \text{\textit{Entropy}},
\end{equation}
where $\mu$ is a small weight applied to entropy regularization. By maximizing entropy, the model encourages a balanced exploration–exploitation trade-off. To balance representation learning and query selection, we enable Stage II every 5 epochs after the 15-epoch Stage \uppercase\expandafter{\romannumeral 1}, with other epochs continuing as Stage \uppercase\expandafter{\romannumeral 1}. This allows periodic refinement of query selection throughout training.

\paragraph{Stage III: Optimal Agents Selection.} After training in the first two stages, the model can effectively identify optimal agents. We then use \( S_Q \) to select agents that maximally enhance the model’s matching for aggregation.

\textbf{\textit{Discussion.}} 
\textit{1. Why not train a separate reward model solely for agent selection?
} 
Because integrating the reward mechanism with the main model allows for more efficient training that aligns closely with the task objective, avoiding extra computational overhead and training complexity. Joint training also reduces reward bias, improves the stability and accuracy of agent selection, and enables the selection strategy to adapt and generalize better, thereby enhancing the overall model’s matching performance.

\textit{2.Why not simply use top-k selection all the time?}
Top-k selection based on scores is rigid and may miss agents with long-term potential but low initial scores. Using Bernoulli sampling increases exploration of such agents, while a soft mask allows flexible contribution weighting, improving overall matching performance. We allow the model to gradually learn an adaptive selection strategy during training, resulting in better cross-modal matching performance.

\textit{3. Why can the Tri-Stage Agents Optimization Strategy select high-quality agents?}
The Tri-Stage Agents Optimization Strategy improves agent quality by focusing on different goals at each training stage: global matching early, local details mid-stage, and a balanced optimization later. Reinforcement learning with joint training enables diverse query exploration and avoids local optima. Dynamic rewards uncover potential queries, while multi-stage, multi-view training improves stability, diversity, and generalization.

\begin{figure}[t]
\centering
\includegraphics[width=\columnwidth]{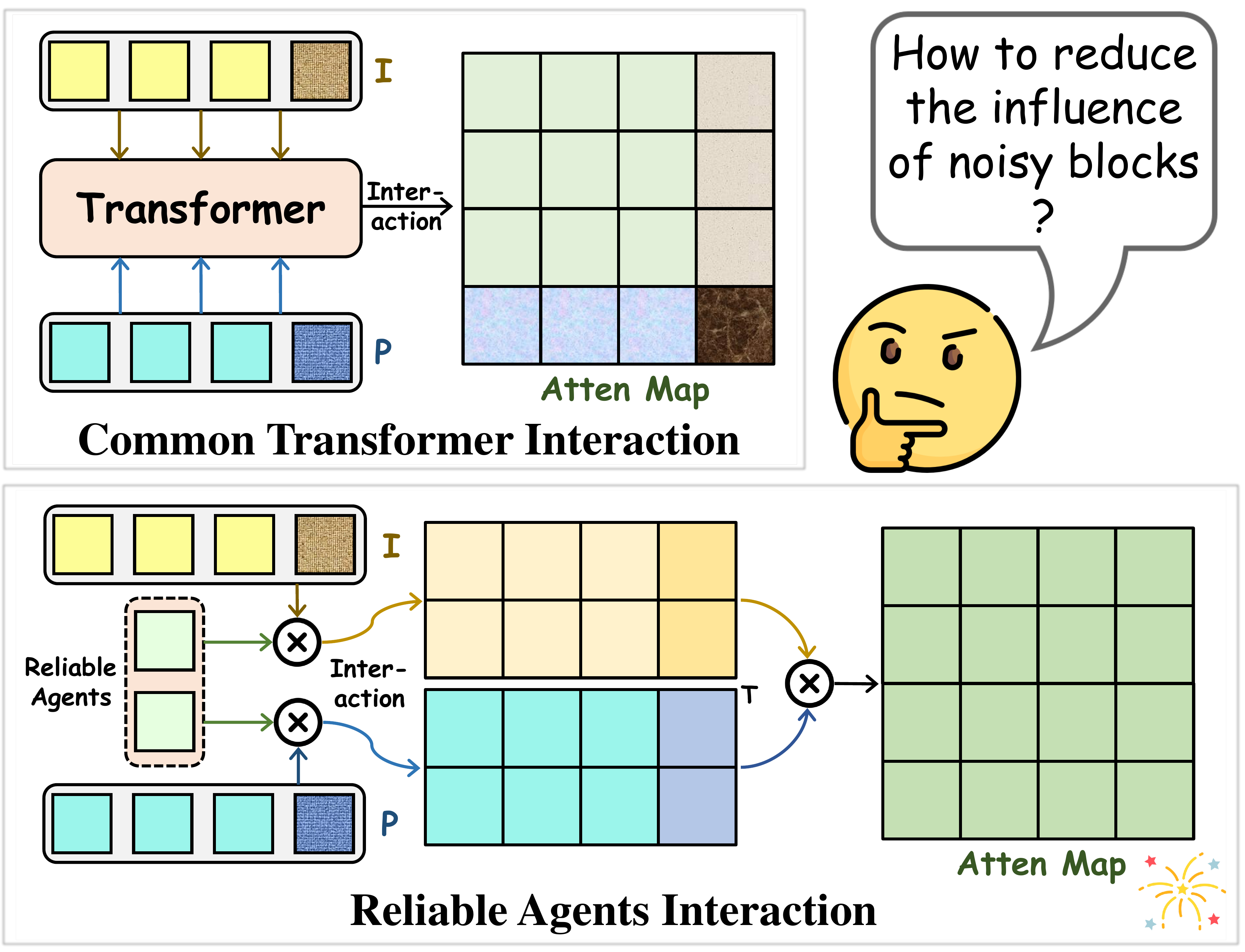} 
\caption{Visualization of difference between common transformer interaction and reliable agents interaction.}
\label{fig3}
\end{figure}

\subsection{Reliable Agents Interaction Module}
After obtaining reliable agents carrying image–point cloud correlation information through the Iterative Agents Selection Module, we leverage these agents as a bridge to refine the transformer aggregation mechanism. As illustrated in Figure 3, standard transformer interaction treats all features equally, often allowing noisy blocks to dominate. In contrast, our reliable agents interaction selects informative features in advance, reducing attention noise and yielding cleaner attention maps with improved feature alignment. By filtering information sources before cross-modal interaction and retaining only high-quality queries for attention computation, the attention becomes more focused, enhancing both the robustness and discriminative power of the learned representations. This strategy effectively suppresses noise from repetitive patterns, non-overlapping regions, and illumination changes, resulting in more accurate image–point cloud matching.

As shown in Figure 2, the reliable agents selected at different stages are denoted as \( A \in \mathbb{R}^{k \times C} \), which are used as queries \( Q \), while the image and point cloud features \( F_i \) and \( F_p \) serve as keys and values. Formally,
\begin{equation}
Q = W^Q \cdot A, 
K_p = W^P \cdot F_p, 
V_p = W^P \cdot F_p, 
\end{equation}
\begin{equation}
K_i = W^I \cdot F_i, \quad V_i = W^I \cdot F_i.
\end{equation}

Here, \( W^Q \), \( W^P \), and \( W^I \) are learnable linear projection matrices that map the input features into a shared embedding space for attention computation.
We then compute the Image-to-Agent Attention (IAA) and Point-to-Agent Attention (PAA) maps as:
\begin{equation}
\operatorname{IAA} = \sigma\left( Q K_p^\top / \sqrt{C} \right),
\operatorname{PAA} = \sigma\left( Q K_i^\top / \sqrt{C} \right),
\end{equation}
\( \sigma(\cdot) \) denote the softmax function. 
These attention maps are then used to aggregate \( V_p \) and \( V_i \), resulting in the fused features \( F_p' \) and \( F_i' \). Patch-level correspondences are established based on feature similarity, followed by dense matching using fine-grained features. Finally, the PnP-RANSAC algorithm is applied to estimate a reliable rigid transformation.

\begin{table*}[t]
\centering
\begin{adjustbox}{width=0.98\textwidth}
\small
\renewcommand{\arraystretch}{1.15}
\setlength{\tabcolsep}{4pt}
\begin{tabular}{lccccc  cccccccc}
\toprule
\cellcolor{gray!8}\textbf{ Dataset} & 
\multicolumn{5}{>{\columncolor{gray!15}}c}{{\textbf{\textcolor{orange!50!black}{RGB-D Scenes v2}}}} & 
\multicolumn{8}{>{\columncolor{gray!25}}c}{{\textbf{\textcolor{blue!30!black}{7-Scenes}}}} \\
\midrule
\cellcolor{gray!8}Model
& \cellcolor{orange!10}Scene-11 
& \cellcolor{orange!10}Scene-12 
& \cellcolor{orange!10}Scene-13 
& \cellcolor{orange!10}Scene-14 
& \cellcolor{orange!10}Mean 
& \cellcolor{cyan!8}Chess 
& \cellcolor{cyan!8}Fire 
& \cellcolor{cyan!8}Heads 
& \cellcolor{cyan!8}Office 
& \cellcolor{cyan!8}Pumpkin 
& \cellcolor{cyan!8}Kitchen 
& \cellcolor{cyan!8}Stairs 
& \cellcolor{cyan!8}Mean \\
\midrule
\cellcolor{gray!8}Mdpt(m) 
& \cellcolor{orange!10}1.74 & \cellcolor{orange!10}1.66 & \cellcolor{orange!10}1.18 & \cellcolor{orange!10}1.39 & \cellcolor{orange!10}1.49 
& \cellcolor{cyan!8}1.78 & \cellcolor{cyan!8}1.55 & \cellcolor{cyan!8}0.80 & \cellcolor{cyan!8}2.03 & \cellcolor{cyan!8}2.25 & \cellcolor{cyan!8}2.13 & \cellcolor{cyan!8}1.84 & \cellcolor{cyan!8}1.49 \\
\midrule
\multicolumn{14}{c}{\textbf{\cellcolor{green!10}\textit{Inlier Ratio} ↑}} \\
\midrule
\cellcolor{gray!8}FCGF-2D3D 
& \cellcolor{orange!10}6.8 & \cellcolor{orange!10}8.5 & \cellcolor{orange!10}11.8 & \cellcolor{orange!10}5.4 & \cellcolor{orange!10}8.1 
& \cellcolor{cyan!8}34.2 & \cellcolor{cyan!8}32.8 & \cellcolor{cyan!8}14.8 & \cellcolor{cyan!8}26 & \cellcolor{cyan!8}23.3 & \cellcolor{cyan!8}22.5 & \cellcolor{cyan!8}6.0 & \cellcolor{cyan!8}22.8 \\
\cellcolor{gray!8}P2-Net 
& \cellcolor{orange!10}9.7 & \cellcolor{orange!10}12.8 & \cellcolor{orange!10}17.0 & \cellcolor{orange!10}9.3 & \cellcolor{orange!10}12.2 
& \cellcolor{cyan!8}55.2 & \cellcolor{cyan!8}46.7 & \cellcolor{cyan!8}13.0 & \cellcolor{cyan!8}36.2 & \cellcolor{cyan!8}32.0 & \cellcolor{cyan!8}32.8 & \cellcolor{cyan!8}5.8 & \cellcolor{cyan!8}31.7 \\
\cellcolor{gray!8}Predator-2D3D 
& \cellcolor{orange!10}17.7 & \cellcolor{orange!10}19.4 & \cellcolor{orange!10}17.2 & \cellcolor{orange!10}8.4 & \cellcolor{orange!10}15.7 
& \cellcolor{cyan!8}34.7 & \cellcolor{cyan!8}33.8 & \cellcolor{cyan!8}16.6 & \cellcolor{cyan!8}25.9 & \cellcolor{cyan!8}23.1 & \cellcolor{cyan!8}22.2 & \cellcolor{cyan!8}7.5 & \cellcolor{cyan!8}23.4 \\
\cellcolor{gray!8}2D3D-MATR 
& \cellcolor{orange!10}\underline{32.8} &\cellcolor{orange!10}\underline{\textbf{34.4}}&\cellcolor{orange!10}\underline{39.2} & \cellcolor{orange!10}\underline{23.3} & \cellcolor{orange!10}\underline{32.4} 
& \cellcolor{cyan!8}\underline{72.1} & \cellcolor{cyan!8}\underline{66.0} & \cellcolor{cyan!8}\underline{31.3} & \cellcolor{cyan!8}\underline{60.7} & \cellcolor{cyan!8}\underline{50.2} & \cellcolor{cyan!8}\underline{\textbf{52.5}} & \cellcolor{cyan!8}\underline{\textbf{18.1}} & \cellcolor{cyan!8}\underline{50.1} \\
\cellcolor{gray!8}B2-3Dnet 
& \cellcolor{orange!10}\textbf{36.4} & \cellcolor{orange!10}32.7 & \cellcolor{orange!10}\textbf{43.8} & \cellcolor{orange!10}\textbf{27.4} & \cellcolor{orange!10}\textbf{35.1}
& \cellcolor{cyan!8}\textbf{73.8} & \cellcolor{cyan!8}\textbf{66.7} & \cellcolor{cyan!8}\textbf{33.1} & \cellcolor{cyan!8}\textbf{61.7} & \cellcolor{cyan!8}\textbf{50.8} & \cellcolor{cyan!8}52.3 & \cellcolor{cyan!8}\textbf{18.1} & \cellcolor{cyan!8}\textbf{50.9} \\
\cellcolor{gray!8}A2SI 
& \cellcolor{orange!10}\textbf{\textcolor{DarkBurntOrange}{40.1}} & \cellcolor{orange!10}\textbf{\textcolor{DarkBurntOrange}{41.1}} & \cellcolor{orange!10}\textbf{\textcolor{DarkBurntOrange}{44.8}} & \cellcolor{orange!10}\textbf{\textcolor{DarkBurntOrange}{28.5}} & \cellcolor{orange!10}\textbf{\textcolor{DarkBurntOrange}{38.6}} 
& \cellcolor{cyan!8}\textbf{\textcolor{DeepTeal}{75.5}} & \cellcolor{cyan!8}\textbf{\textcolor{DeepTeal}{68.9}} & \cellcolor{cyan!8}\textbf{\textcolor{DeepTeal}{40.0}} & \cellcolor{cyan!8}\textbf{\textcolor{DeepTeal}{66.6}} & \cellcolor{cyan!8}\textbf{\textcolor{DeepTeal}{53.6}} & \cellcolor{cyan!8}\textbf{\textcolor{DeepTeal}{55.6}} & \cellcolor{cyan!8}\textbf{\textcolor{DeepTeal}{18.2}} & \cellcolor{cyan!8}\textbf{\textcolor{DeepTeal}{54.1}} \\
\midrule
\multicolumn{14}{c}{\textbf{\cellcolor{green!10}\textit{Feature Matching Recall} ↑}} \\
\midrule
\cellcolor{gray!8}FCGF-2D3D 
& \cellcolor{orange!10}11.1 & \cellcolor{orange!10}30.4 & \cellcolor{orange!10}51.5 & \cellcolor{orange!10}15.5 & \cellcolor{orange!10}27.1 
& \cellcolor{cyan!8}\textbf{99.7} & \cellcolor{cyan!8}98.2 & \cellcolor{cyan!8}69.9 & \cellcolor{cyan!8}97.1 & \cellcolor{cyan!8}83.0 & \cellcolor{cyan!8}87.7 & \cellcolor{cyan!8}16.2 & \cellcolor{cyan!8}78.8 \\
\cellcolor{gray!8}P2-Net 
& \cellcolor{orange!10}48.6 & \cellcolor{orange!10}65.7 & \cellcolor{orange!10}82.5 & \cellcolor{orange!10}41.6 & \cellcolor{orange!10}59.6  
& \cellcolor{cyan!8}\textbf{\textcolor{DeepTeal}{100.0}} & \cellcolor{cyan!8}99.3 & \cellcolor{cyan!8}58.9 & \cellcolor{cyan!8}\textbf{99.1} & \cellcolor{cyan!8}87.2 & \cellcolor{cyan!8}92.2 & \cellcolor{cyan!8}16.2 & \cellcolor{cyan!8}79 \\
\cellcolor{gray!8}Predator-2D3D 
& \cellcolor{orange!10}86.1 & \cellcolor{orange!10}89.2 & \cellcolor{orange!10}63.9 & \cellcolor{orange!10}24.3 & \cellcolor{orange!10}65.9 
& \cellcolor{cyan!8}91.3 & \cellcolor{cyan!8}95.1 & \cellcolor{cyan!8}76.6 & \cellcolor{cyan!8}88.6 & \cellcolor{cyan!8}79.2 & \cellcolor{cyan!8}80.6 & \cellcolor{cyan!8}31.1 & \cellcolor{cyan!8}77.5 \\
\cellcolor{gray!8}2D3D-MATR 
& \cellcolor{orange!10}\underline{\textbf{98.6}} & \cellcolor{orange!10}\underline{98.0} & \cellcolor{orange!10}\underline{\textbf{88.7}} & \cellcolor{orange!10}\underline{77.9} & \cellcolor{orange!10}\underline{90.8} 
& \cellcolor{cyan!8}\underline{\textbf{\textcolor{DeepTeal}{100.0}}} & \cellcolor{cyan!8}\underline{99.6} & \cellcolor{cyan!8}\underline{\textbf{98.6}} & \cellcolor{cyan!8}\underline{\textbf{\textcolor{DeepTeal}{100.0}}} & \cellcolor{cyan!8}\underline{\textbf{92.4}} & \cellcolor{cyan!8}\underline{\textbf{95.9}} & \cellcolor{cyan!8}\underline{58.2} & \cellcolor{cyan!8}\underline{\textbf{92.1}} \\
\cellcolor{gray!8}B2-3Dnet 
& \cellcolor{orange!10}\textbf{\textcolor{DarkBurntOrange}{100.0}} & \cellcolor{orange!10}\textbf{99.0} & \cellcolor{orange!10}\textbf{\textcolor{DarkBurntOrange}{92.8}} & \cellcolor{orange!10}\textbf{\textcolor{DarkBurntOrange}{85.8}} & \cellcolor{orange!10}\textbf{\textcolor{DarkBurntOrange}{94.4}}
& \cellcolor{cyan!8}\textbf{\textcolor{DeepTeal}{100.0}} & \cellcolor{cyan!8}\textbf{\textcolor{DeepTeal}{100.0}} & \cellcolor{cyan!8}\textbf{98.6} & \cellcolor{cyan!8}\textbf{\textcolor{DeepTeal}{100.0}} & \cellcolor{cyan!8}\textbf{\textcolor{DeepTeal}{92.7}} & \cellcolor{cyan!8}95.6 & \cellcolor{cyan!8}\textbf{64.9} & \cellcolor{cyan!8}\textbf{\textcolor{DeepTeal}{93.1}} \\
\cellcolor{gray!8}A2SI
& \cellcolor{orange!10}\textbf{98.6} & \cellcolor{orange!10}\textbf{\textcolor{DarkBurntOrange}{100.0}} & \cellcolor{orange!10}\textbf{\textcolor{DarkBurntOrange}{92.8}} & \cellcolor{orange!10}\textbf{85.6} & \cellcolor{orange!10}\textbf{94.3} 
& \cellcolor{cyan!8}\textbf{\textcolor{DeepTeal}{100.0}} & \cellcolor{cyan!8}\textbf{99.8} & \cellcolor{cyan!8}\textbf{\textcolor{DeepTeal}{100.0}} & \cellcolor{cyan!8}\textbf{\textcolor{DeepTeal}{100.0}} & \cellcolor{cyan!8}90.6 & \cellcolor{cyan!8}\textbf{\textcolor{DeepTeal}{96.0}} & \cellcolor{cyan!8}\textbf{\textcolor{DeepTeal}{65.1}} & \cellcolor{cyan!8}\textbf{\textcolor{DeepTeal}{93.1}} \\
\midrule
\multicolumn{14}{c}{\textbf{\cellcolor{green!10}\textit{Registration Recall} ↑}} \\
\midrule
\cellcolor{gray!8}FCGF-2D3D 
& \cellcolor{orange!10}26.5 & \cellcolor{orange!10}41.2 & \cellcolor{orange!10}37.1 & \cellcolor{orange!10}16.8 & \cellcolor{orange!10}30.4 
& \cellcolor{cyan!8}89.5 & \cellcolor{cyan!8}79.7 & \cellcolor{cyan!8}19.2 & \cellcolor{cyan!8}85.9 & \cellcolor{cyan!8}69.4 & \cellcolor{cyan!8}79.0 & \cellcolor{cyan!8}6.8 & \cellcolor{cyan!8}61.4 \\
\cellcolor{gray!8}P2-Net 
& \cellcolor{orange!10}40.3 & \cellcolor{orange!10}40.2 & \cellcolor{orange!10}41.2 & \cellcolor{orange!10}31.9 & \cellcolor{orange!10}38.4  
& \cellcolor{cyan!8}96.9 & \cellcolor{cyan!8}86.5 & \cellcolor{cyan!8}20.5 & \cellcolor{cyan!8}91.7 & \cellcolor{cyan!8}75.3 & \cellcolor{cyan!8}85.2 & \cellcolor{cyan!8}4.1 & \cellcolor{cyan!8}65.7 \\
\cellcolor{gray!8}Predator-2D3D 
& \cellcolor{orange!10}44.4 & \cellcolor{orange!10}41.2 & \cellcolor{orange!10}21.6 & \cellcolor{orange!10}13.7 & \cellcolor{orange!10}30.2 
& \cellcolor{cyan!8}69.6 & \cellcolor{cyan!8}60.7 & \cellcolor{cyan!8}17.8 & \cellcolor{cyan!8}62.9 & \cellcolor{cyan!8}56.2 & \cellcolor{cyan!8}62.6 & \cellcolor{cyan!8}9.5 & \cellcolor{cyan!8}48.5 \\
\cellcolor{gray!8}2D3D-MATR 
& \cellcolor{orange!10}\underline{\textbf{63.9}} & \cellcolor{orange!10}\underline{53.9} & \cellcolor{orange!10}\underline{58.8} & \cellcolor{orange!10}\underline{49.1} & \cellcolor{orange!10}\underline{56.4} 
& \cellcolor{cyan!8}\underline{96.9} & \cellcolor{cyan!8}\underline{90.7} & \cellcolor{cyan!8}\underline{52.1} & \cellcolor{cyan!8}\underline{95.5} & \cellcolor{cyan!8}\underline{80.9} & \cellcolor{cyan!8}\underline{86.1} & \cellcolor{cyan!8}\underline{28.4} & \cellcolor{cyan!8}\underline{75.8} \\
\cellcolor{gray!8}B2-3Dnet 
& \cellcolor{orange!10}58.3 & \cellcolor{orange!10}\textbf{60.8} & \cellcolor{orange!10}\textbf{74.2} & \cellcolor{orange!10}\textbf{\textcolor{DarkBurntOrange}{60.2}} & \cellcolor{orange!10}\textbf{63.4}
& \cellcolor{cyan!8}\textbf{\textcolor{DeepTeal}{98.3}} & \cellcolor{cyan!8}\textbf{90.5} & \cellcolor{cyan!8}\textbf{56.2} & \cellcolor{cyan!8}\textbf{96.4} & \cellcolor{cyan!8}\textbf{\textcolor{DeepTeal}{84.0}} & \cellcolor{cyan!8}\textbf{86.1} & \cellcolor{cyan!8}\textbf{\textcolor{DeepTeal}{32.4}} & \cellcolor{cyan!8}\textbf{77.7} \\ 
\cellcolor{gray!8}A2SI
& \cellcolor{orange!10}\textbf{\textcolor{DarkBurntOrange}{69.4}} & \cellcolor{orange!10}\textbf{\textcolor{DarkBurntOrange}{76.5}} & \cellcolor{orange!10}\textbf{\textcolor{DarkBurntOrange}{87.6}} & \cellcolor{orange!10}\textbf{58.8} & \cellcolor{orange!10}\textbf{\textcolor{DarkBurntOrange}{73.1}} 
& \cellcolor{cyan!8}\textbf{97.6} & \cellcolor{cyan!8}\textbf{\textcolor{DeepTeal}{95.4}} & \cellcolor{cyan!8}\textbf{\textcolor{DeepTeal}{68.5}} & \cellcolor{cyan!8}\textbf{\textcolor{DeepTeal}{98.0}} & \cellcolor{cyan!8}\textbf{81.2} & \cellcolor{cyan!8}\textbf{\textcolor{DeepTeal}{90.5}} & \cellcolor{cyan!8}\textbf{28.5} & \cellcolor{cyan!8}\textbf{\textcolor{DeepTeal}{79.9}} \\
\bottomrule
\end{tabular}
\end{adjustbox}
\caption{Evaluation results on RGB-D Scenes v2 and 7-Scenes. \textbf{\textcolor{DarkBurntOrange}{Orange}} and \textbf{\textcolor{DeepTeal}{Blue}} numbers highlight the best, the second best are \textbf{Boldfaced} and the baseline are \underline{underlined}.}
\label{tab:RTERRE}
\end{table*}

\subsection{Model Training \& Post-Processing Details}
Let us examine the loss functions for the coarse and fine-matching networks. Both \( \mathcal{L}_{\text{coarse}} \) and \( \mathcal{L}_{\text{fine}} \) utilize a general circle loss \cite{circleloss,circleloss1}. For a given anchor descriptor \( d_t \), the descriptors of its positive and negative pairs are represented as \( \mathcal{D}_t^P \) and \( \mathcal{D}_t^N \), respectively. The loss function is defined as follows:
\begin{equation}
\begin{aligned}
\mathcal{L}_t = \frac{1}{\gamma} \log\biggl[1 + \Bigl(&\sum_{d^j \in \mathcal{D}^P_t} e^{\beta^{t,j}_p (d^j_t - \Delta_p)}\Bigr) \\
&\cdot \Bigl(\sum_{d^k \in \mathcal{D}^N_t} e^{\beta^{t,k}_n (\Delta_n - d^k_t)}\Bigr)\biggr].
\end{aligned}
\end{equation}

The Tri-Stage Agents Optimization strategy is applied only during training. The overall loss function is defined as:
\begin{equation}
\mathcal{L} = \lambda_1 \cdot \mathcal{L}_t + \lambda_2 \cdot \mathcal{L}_{full},
\end{equation}
where \( \mathcal{L}_{full} \) is activated every 5 epochs during Stage~\uppercase\expandafter{\romannumeral2}.

\section{Experiments}
\subsection{Datasets and Implementation Details}
Based on the 2D3D-MATR \cite{matr2d3d} benchmark, we conduct extensive experiments and ablation studies on two challenging benchmarks: RGB-D Scenes v2 \cite{rgbdv2} and 7Scenes \cite{7scenes}.

\textbf{Dataset.} \textit{RGB-D Scenes V2} consists of 14 scenes containing furniture. For each scene, we create point cloud fragments from every 25 consecutive depth frames and sample one RGB image per 25 frames. We select image-point-cloud pairs with an overlap ratio of at least 30\%. Scenes 1-8 are used for training, 9-10 for validation, and 11-14 for testing, resulting in 1,748 training pairs, 236 validation pairs, and 497 testing pairs.

The \textit{7-Scenes} is a collection of tracked RGB-D camera frames. All seven indoor scenes were recorded from a handheld Kinect RGB-D camera at 640×480 resolution. We select image-to-point-cloud pairs from each scene with at least 50\% overlap, adhering to the official sequence split for training, validation, and testing. This results in 4,048 training pairs, 1,011 validation pairs, and 2,304 testing pairs.

\textbf{Implementation Details.} We used an NVIDIA Geforce RTX 3090 GPU for training. We implement our model using
PyTorch 1.13.1.
We set the number of attention layers to \( n = 3 \) by default.  
The initial $\tau$ is set to $20.0$ and decayed by a factor of $0.9$ every $10$ epochs until reaching a minimum of $5.0$.  
We use an entropy regularization weight of $\mu = 0.01$ and set the loss balancing factors to $\lambda_1 = \lambda_2 = 1$.  
The number of agents is fixed at $12$, which also matches the number of queries selected by the top-$k$ strategy.  
Additionally, we set the soft masking coefficient to $\beta = 0.3$ to stabilize gradient propagation.

\textbf{Metrics.} We evaluate models with three metrics: Inlier Ratio (IR) — the percentage of pixel-point matches with a 3D distance below 5 cm; Feature Matching Recall (FMR) — the percentage of image–point-cloud pairs with an inlier ratio above 10\%; and Registration Recall (RR) — the percentage of image–point-cloud pairs with an RMSE below 10 cm. Patch Inlier Ratio (PIR) evaluates coarse-level alignment by measuring the fraction of patch correspondences with sufficient overlap under GT transformation.

\begin{figure*}[t]
\centering
\includegraphics[width=\textwidth]{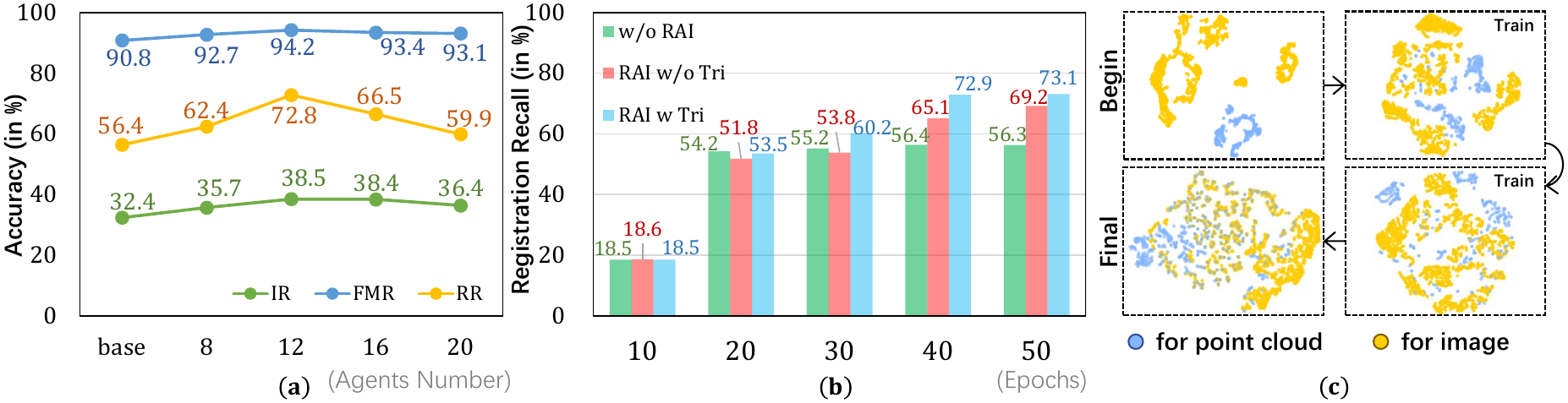} 
\caption{Visualization of difference in (a) agents number (b) Convergence (c) t-sne.}
\label{fig7}
\end{figure*}

\begin{table}[!b]
\centering
\setlength{\tabcolsep}{1mm} 
\renewcommand{\arraystretch}{0.9} 
\resizebox{\columnwidth}{!}{
\begin{tabular}{c||cccc||cccc}
\toprule
Method & Phase & RAI & Tri & Topk & PIR & IR & FMR & RR \\
\midrule
M1 &   &   &   &   & 48.5 & 32.5 & 91.0 & 56.4 \\
M2 & \checkmark &   &   &  & 53.4 & 34.5 & 91.6 & 57.5 \\
M3 &    & \checkmark &   &   & 56.8 & 37.6 & 92.8 & 68.4 \\
M4 &  & \checkmark  &   &  \checkmark & 58.9 & 38.2 & 93.4 & 70.1 \\
M5 &   &  \checkmark & \checkmark &   & \underline{60.2} & \underline{38.5} & \underline{94.2} & \underline{72.8} \\
M6 & \checkmark  & \checkmark  &  &   & 56.1 & 37.8 & 93.1 & 69.1 \\
M7 &  \checkmark &  \checkmark &   & \checkmark & 59.8 & 38.3 & 93.7 & 70.2 \\
M8 & \checkmark & \checkmark & \checkmark & & \textbf{\textcolor{DarkBurntOrange}{60.8}} & \textbf{\textcolor{DarkBurntOrange}{38.6}} & \textbf{\textcolor{DarkBurntOrange}{94.3}} & \textbf{\textcolor{DarkBurntOrange}{73.1}} \\
\bottomrule
\end{tabular}}
\caption{Ablation study results on RGB-D Scenes v2. Phase indicates the use of phase map enhancement, RAI denotes the Reliable Agents Interaction module, Tri refers to the Tri-Stage Agents Optimization, and Topk represents the Top-$k$ selection strategy.
\textbf{\textcolor{DarkBurntOrange}{Orange}} numbers highlight the best, the second best are \underline{underlined}.}
\label{tab:ablation}
\end{table}

\subsection{Evaluations on Dataset}
We compare our approach with 2D3D-MATR~\cite{matr2d3d} and other baselines~\cite{fcgf2d3d,p2,predator2d3d,cheng11} on the RGB-D Scenes v2 and 7-Scenes datasets (Table 1).  

On the RGB-D Scenes v2 dataset, our method shows significant improvements in challenging, narrow indoor environments such as Scene-11, Scene-12, and Scene-13, where geometric ambiguities often cause mismatches.  
By leveraging edge information extracted from the IAS phase map and focusing on critical regions, our RAI module aggregates information through reliable agents, leading to more robust correspondence establishment.  
As a result, we achieve a 28.8 percentage point (pp) gain in registration recall over 2D3D-MATR on Scene-12 (from 58.8\% to 87.6\%).  
Overall, our method improves the mean inlier ratio by 6.2 pp (from 32.4\% to 38.6\%), the mean feature matching recall by 3.5 pp (from 90.8\% to 94.3\%), and boosts the mean registration recall by 16.7 pp (from 56.4\% to 73.1\%).

On the 7-Scenes dataset, which features larger scale variations, our method consistently outperforms previous approaches, achieving a mean registration recall of 79.9\%, surpassing 2D3D-MATR by 4.4 pp (from 75.5\%).  
Although in the Stairs scene — characterized by repetitive textures and lacking distinctive structural regions — the advantage of our method is limited, notable improvements are observed in challenging cases such as Heads and Kitchen, with registration recall gains of 16.6 pp and 4.4 pp over 2D3D-MATR, respectively.  
These results demonstrate the robustness and generalization ability of our approach across diverse indoor environments.

\subsection{Ablation Studies}
To evaluate the contribution of each module to the overall performance, we conduct an ablation study on the RGB-D Scenes v2 dataset, as shown in Table 2.
Using the phase map (M2) improves registration recall by 1.1\% over the baseline.  
Adding the RAI module (M3) further boosts performance across all metrics. 
The Tri-Stage Agents Optimization (M4) significantly boosts registration recall by progressively refining correspondences through a multi-stage selection and optimization process. This strategy proves critical for accurate alignment, as it adaptively filters and enhances informative queries across stages.
In comparison, the Top-$k$ strategy (M5) brings moderate improvements by selecting reliable queries from a larger pool of candidates. However, its single-stage, score-based selection lacks the refinement capability of the Tri-Stage Agents Optimization, making it less effective in capturing high-quality correspondences.
M6 replaces adaptive query selection with 12 fixed learnable queries, showing that selecting from a larger pool of queries benefits registration performance.
M7 adopts a simple Top-$k$ selection from redundant learnable queries, outperforming fixed queries in M6 and highlighting the advantage of selecting reliable queries.
Our full model (M8) employs the proposed Tri-Stage Agents Optimization to select agents, further improving the upper bound of performance and validating the complementary benefits of all modules.

\begin{figure*}[t]
\centering
\includegraphics[width=\textwidth]{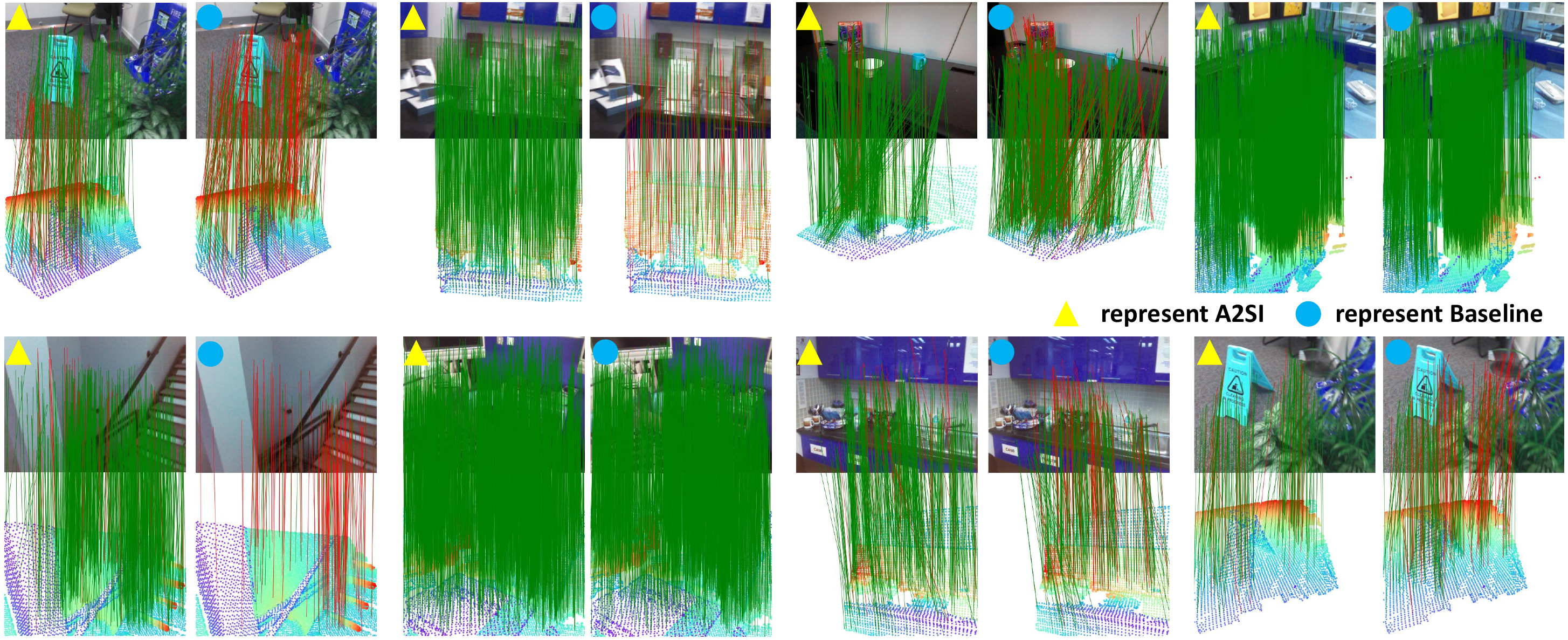} 
\caption{Visualization of matching accuracy: correct (green) if error $< 60$ px; otherwise red.}
\label{fig4}
\end{figure*}

\begin{figure}[!b]
\centering
\includegraphics[width=\columnwidth]{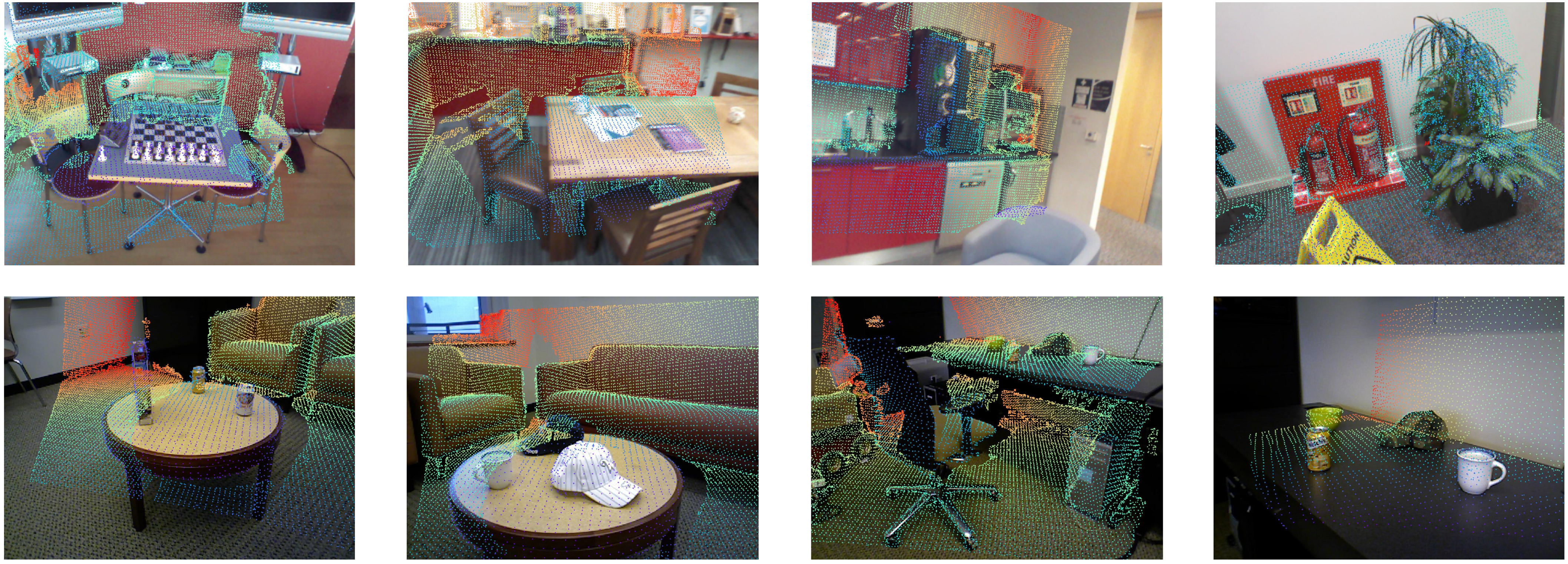} 
\caption{Visualization of A2SI.}
\label{fig5}
\end{figure}

As shown in Figure 4(a), the ablation results indicate that using 12 agents achieves the best registration recall and inlier ratio. Increasing the number of agents beyond this slightly degrades performance, likely due to redundancy.
In Figure 4(b), our method with both RAI and Tri not only attains higher registration recall but also converges faster than the variants without them, highlighting both its effectiveness and training efficiency.
Finally, Figure 4(c) shows that the modality gap between image and point cloud features gradually narrows during training, resulting in more consistent cross-modal representations.

\subsection{Qualitative Results}
Although we have conducted extensive experiments, we further perform comprehensive visualization analyses to intuitively demonstrate the model’s performance.

In Figure 5, a match is regarded as correct if the projected point cloud keypoint falls within 30 pixels of its corresponding image point, allowing us to distinguish correct matches (green) from incorrect ones. The yellow triangles denote A2SI, while the blue circles represent the baseline 2D3D-MATR.  
It can be observed that our method, benefiting from better feature aggregation and the incorporation of edge information, significantly improves matching performance in challenging scenes — as reflected by the increased number of green lines and the reduced number of red ones.
In Figure 6, we project the point cloud onto the image, where the matches on structural objects such as tables, cabinets, and chairs appear highly accurate.
This result demonstrates the model’s ability to capture reliable correspondences on meaningful geometric structures, which are critical for robust registration in indoor environments.

\begin{figure}[t]
\centering
\includegraphics[width=\columnwidth]{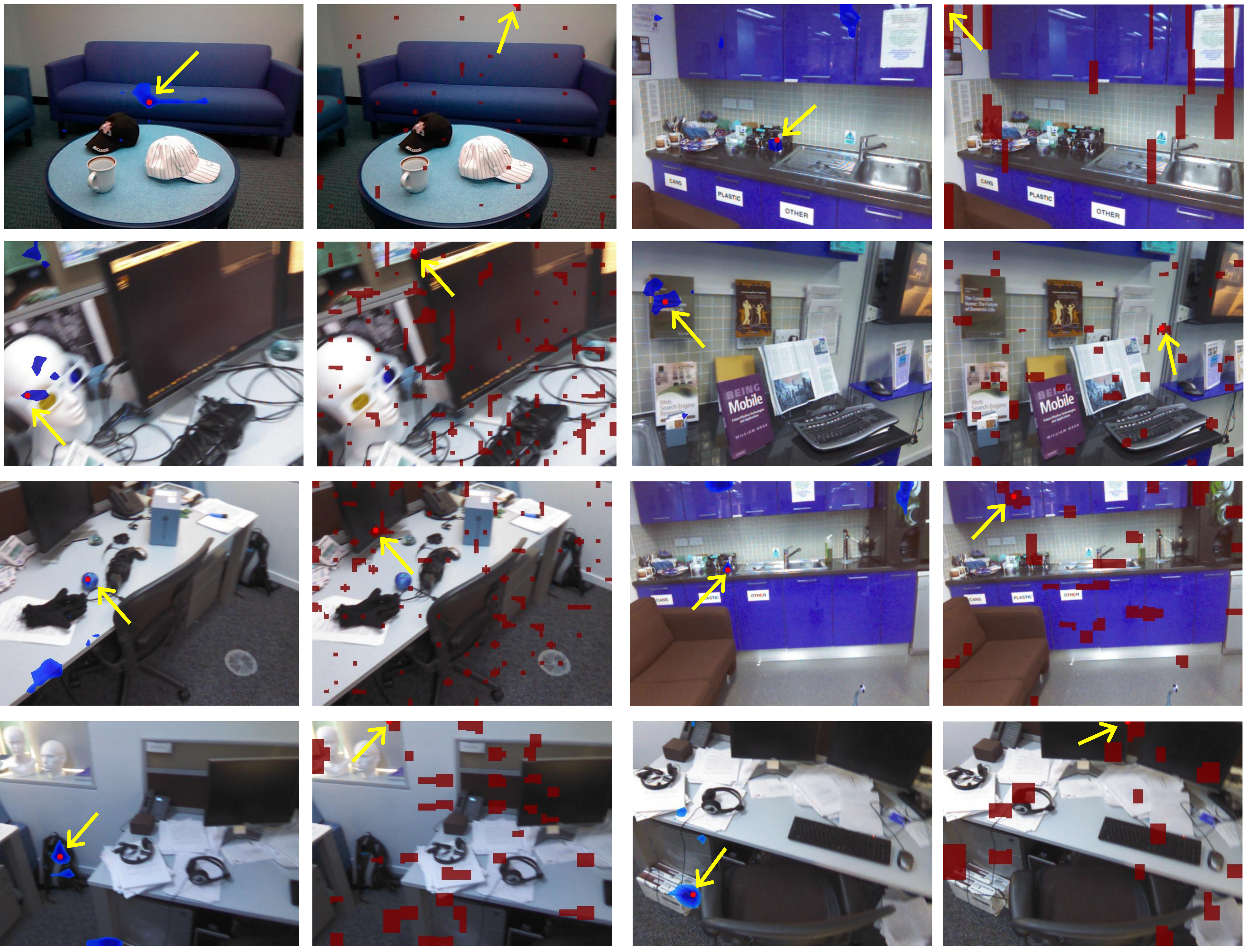} 
\caption{Visualization of differences in attention focus.}
\label{fig6}
\end{figure}

In Figure 7, we visualize the attention regions from the last layer of the transformer. The red dots indicate the highest attention points, while the red and blue regions highlight the top 5\% of attention scores.
As shown in the figure, thanks to reliable information aggregation, our method focuses more on key regions such as distinct objects, rather than distributing attention broadly and uniformly like conventional transformers.
This targeted attention allows the model to extract more meaningful features for registration, improving both robustness and matching accuracy — especially in cluttered or complex scenes where precise correspondence is essential.

\section{Conclusion}
In this paper, we introduce a novel Adaptive Agent Selection and Interaction (A2SI) Network for image-to-point cloud registration.  
Our method enhances features by incorporating edge information from phase maps and adopts reinforcement learning-inspired strategies to select informative agents.  
We further aggregate image and point cloud features through reliable agents, which improves registration accuracy and reduces computational complexity.  
Extensive experiments on the RGB-D Scenes v2 and 7-Scenes datasets demonstrate that our A2SI approach surpasses existing methods in image-to-point cloud registration.

\bibliography{aaai2026}
\makeatletter
\@ifundefined{isChecklistMainFile}{
  \newif\ifreproStandalone
  \reproStandalonetrue
}{
  \newif\ifreproStandalone
  \reproStandalonefalse
}
\makeatother

\setlength{\leftmargini}{20pt}
\makeatletter\def\@listi{\leftmargin\leftmargini \topsep .5em \parsep .5em \itemsep .5em}
\def\@listii{\leftmargin\leftmarginii \labelwidth\leftmarginii \advance\labelwidth-\labelsep \topsep .4em \parsep .4em \itemsep .4em}
\def\@listiii{\leftmargin\leftmarginiii \labelwidth\leftmarginiii \advance\labelwidth-\labelsep \topsep .4em \parsep .4em \itemsep .4em}\makeatother

\setcounter{secnumdepth}{0}
\renewcommand\thesubsection{\arabic{subsection}}
\renewcommand\labelenumi{\thesubsection.\arabic{enumi}}

\newcounter{checksubsection}
\newcounter{checkitem}[checksubsection]

\newcommand{\checksubsection}[1]{%
  \refstepcounter{checksubsection}%
  \paragraph{\arabic{checksubsection}. #1}%
  \setcounter{checkitem}{0}%
}

\newcommand{\checkitem}{%
  \refstepcounter{checkitem}%
  \item[\arabic{checksubsection}.\arabic{checkitem}.]%
}
\newcommand{\question}[2]{\normalcolor\checkitem #1 #2 \color{blue}}
\newcommand{\ifyespoints}[1]{\makebox[0pt][l]{\hspace{-15pt}\normalcolor #1}}

\ifreproStandalone
\end{document}